%% file: vcmm.tex
\documentclass{article}
\usepackage{spconf,amsmath,amssymb,graphicx,booktabs,multirow,hyperref}
\usepackage{dblfloatfix} 
\usepackage[table]{xcolor} 

\ifdefined\XeTeXrevision

\fi

\usepackage{hyperref}
\hypersetup{
colorlinks=true,
linkcolor=red,
filecolor=red,
urlcolor=magenta,
citecolor= blue 
}

\title{VCMM: Variance-Calibrated Momentum for Multimodal Learning}

\name{
\textit{Zhongjing Gu},
\textit{Chenyang Huang},
\textit{Yufa Feng},
\textit{Chong He},
\textit{Qinxu Ding},
\textit{Yiming Cui}
}

\address{}

\begin{document}
\ninept
\maketitle

\begin{abstract}
Multimodal joint training often suffers from modality imbalance, where a dominant modality suppresses the optimization of others. Existing methods mainly balance modality learning by modulating gradient magnitudes or directions, modifying optimization objectives, or adjusting training strategies, with most interventions focusing on the current update. However, when combined with widely used momentum-based optimizers, the update also incorporates accumulated information from previous gradients, which is not explicitly addressed by current-step modulation alone. To address this issue, we propose \textbf{V}ariance-\textbf{C}alibrated \textbf{M}omentu\textbf{M} \textbf{(VCMM)}, which adapts gradient memory to modality-specific gradient dynamics. Specifically, VCMM estimates minibatch noise and temporal drift online and uses their relative strength to determine modality-specific momentum through a Kalman-inspired controller. We further center the control signal across modalities and apply exact bias correction for the time-varying first moment, enabling adaptive gradient memory without extra network passes or explicit learning-rate scaling. Experiments on four multimodal benchmarks demonstrate consistent improvements with modest training overhead.
\end{abstract}


\section{Introduction}
\label{sec:intro}

Multimodal learning integrates complementary information from different
modalities~\cite{ngiam2011multimodal,baltrusaitis2019multimodal,huang2021better}, but joint training often suffers from optimization imbalance~\cite{wang2020hard,peng2022ogm,wu2022greedy,zhou2025selfpaced}.
A dominant modality can suppress the learning of others, leading to
under-optimized unimodal representations and limiting the benefit of
multimodal fusion~\cite{wang2020hard,wu2022greedy,huang2022competition}.

Existing methods for modality imbalance mainly intervene through \emph{gradient modulation}, \emph{objective optimization}, or \emph{training strategy adjustment}. Gradient-based methods regulate modality-wise gradient magnitudes~\cite{peng2022ogm,li2023agm}, with recent work further considering gradient directions~\cite{guo2024cggm}. Objective-based methods coordinate multimodal and unimodal learning objectives~\cite{wei2024mmpareto}, while training-based methods adjust modality update schedules or sample sequences~\cite{hua2024reconboost,guan2025bss}. Although these methods adopt different balancing mechanisms, most of them primarily regulate the current optimization step. Indeed, with momentum-based optimizers, the current update also incorporates gradient information accumulated from previous steps~\cite{sutskever2013momentum,gitman2019momentum,zhang2019yellowfin}. This cross-step accumulation is typically governed by a shared momentum coefficient across modalities, implicitly assuming that different modalities require the same \emph{gradient memory}.

\begin{figure}[!t]
    \centering
    \includegraphics[width=\columnwidth]{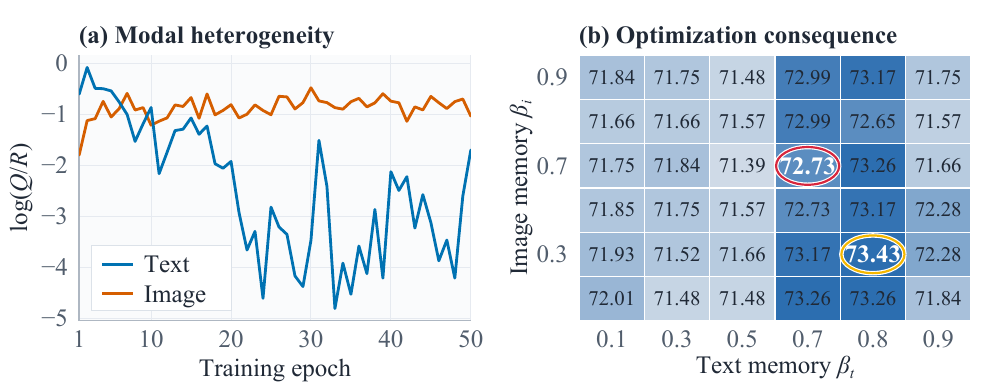}
    \caption{
(a) Drift-to-noise ratios $Q/R$ of Image and Text.
(b) Performance under shared and asymmetric momentum settings.  }
    \label{fig:motivation}
\end{figure}

However, the assumption of shared gradient memory across modalities may not always hold. Momentum governs the trade-off between suppressing minibatch noise and tracking temporal gradient changes~\cite{yao2026sgdf}. A larger momentum provides stronger smoothing but responds more slowly to gradient drift, whereas a smaller momentum tracks gradient changes faster but retains more stochastic noise. Therefore, the appropriate gradient memory should depend on the relative strength of \emph{minibatch noise} and \emph{temporal drift}. As illustrated in Fig.~\ref{fig:motivation}, we characterize this relation using the drift-to-noise ratio $Q/R$, where $Q$ and $R$ denote temporal drift and minibatch noise, respectively. Figures~\ref{fig:motivation} (a) show that different modalities exhibit distinct $Q/R$ dynamics and that these differences evolve throughout training. Furthermore, Fig.~\ref{fig:motivation} (b) shows that the best asymmetric momentum setting outperforms the best shared setting. These observations suggest that a shared momentum coefficient may be suboptimal for multimodal joint training. Therefore, different modalities require distinct and time-varying gradient memory.

To this end, we propose
\textbf{V}ariance-\textbf{C}alibrated \textbf{M}omentu\textbf{M} \textbf{(VCMM)}, which adapts gradient memory to modality-specific gradient dynamics.
Specifically, VCMM models each minibatch gradient as a noisy observation of a time-varying latent gradient~\cite{vuckovic2018kalman} and estimates the minibatch noise and temporal drift of each modality online.
The resulting drift-to-noise ratio is then mapped to a modality-specific momentum coefficient through a Kalman-inspired controller~\cite{kalman1960new}.
We further center the control signal and apply exact bias correction to the time-varying first moment, preserving relative modal differences while anchoring the control signal to the base momentum and correcting initialization bias.
Together, these designs enable VCMM to adapt modality-specific gradient memory without extra network passes or modality-specific learning-rate scaling. Experiments on four benchmarks show consistent improvements over competitive baselines.
\begin{figure*}[!t]
    \centering
    \vspace{-2em}
    \includegraphics[width=\textwidth]{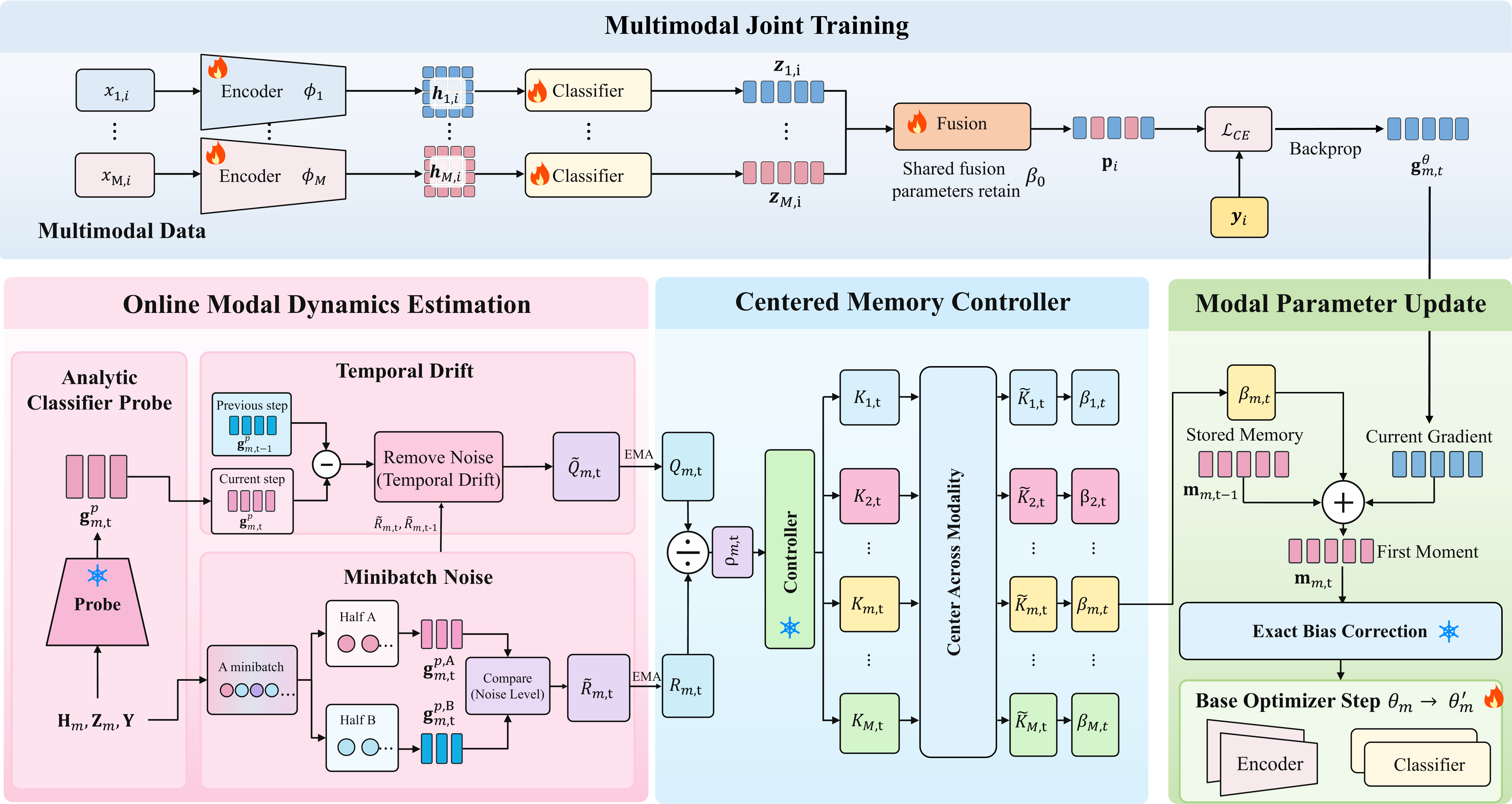}
    \vspace{-2em}
    \caption{
    Overview of Variance-Calibrated MomentuM (VCMM). VCMM estimates modality-wise gradient dynamics with a lightweight probe and converts them into modality-specific momentum through a centered Kalman-inspired controller for adaptive gradient memory.
    }
    \label{fig:vcmm_architecture}
\end{figure*}






\section{Variance-Calibrated Momentum}
\label{sec:method}
In this section, we present VCMM, whose overall framework is illustrated in Figure~\ref{fig:vcmm_architecture}.
We first introduce the multimodal learning setting in
Section~\ref{sec:multimodal_learning}, and in Section~\ref{sec:online_dynamics},
we estimate modality-specific minibatch noise and temporal drift to
characterize the gradient dynamics of different modalities. Based on these
estimates, we introduce the centered memory controller in
Section~\ref{sec:memory_controller}, which converts modal dynamics into
modality-specific momentum coefficients. Finally, in
Section~\ref{sec:modal_update}, we incorporate the resulting momentum into
multimodal joint optimization with bias correction.

\subsection{Multimodal Learning}
\label{sec:multimodal_learning}

We consider a multimodal classification task with $M$ modalities. Let
$\mathcal{D}=\{(\mathbf x_{1,i},\dots,\mathbf x_{M,i},
\mathbf y_i)\}_{i=1}^{N}$ denote the training
set, where $\mathbf x_{m,i}$ is the input of modality $m$ and
$\mathbf y_i\in\{0,1\}^{C}$ is the one-hot label over $C$ classes.
Each modality is
processed by an encoder $\phi_m(\cdot;\boldsymbol{\theta}_m)$,
yielding the representation
$\mathbf h_{m,i}=\phi_m(\mathbf x_{m,i};\boldsymbol{\theta}_m)$
and modal logits $\mathbf z_{m,i}$. The modal
predictions are fused as
$\mathbf p_i=\operatorname{Fusion}(\mathbf z_{1,i},\dots,\mathbf z_{M,i})$.
The model parameters $\boldsymbol{\Theta}$ are jointly optimized
using the cross-entropy objective:
\begin{equation}
\mathcal{L}(\boldsymbol{\Theta})
=
-\frac{1}{N}\sum_{i=1}^{N}
\mathbf y_i^{\top}\log \mathbf p_i.
\label{eq:objective}
\end{equation}

\subsection{Online Modal Dynamics Estimation}
\label{sec:online_dynamics}

The appropriate gradient memory of each modality depends on the relative
strength of stochastic noise and temporal gradient variation. Therefore,
the key challenge is to estimate these two dynamics online during joint
training. Directly collecting gradient statistics over all modality-specific
parameters, however, would introduce considerable computational overhead.
To make this estimation practical, we construct a lightweight classifier
probe that reuses the representations and logits from the original forward
pass.

\noindent\textbf{Analytic Classifier Probe.}
At iteration $t$, $\mathbf H_m$, $\mathbf Z_m$, and $\mathbf Y$ stack
$\mathbf h_{m,j}^{\top}$, $\mathbf z_{m,j}^{\top}$, and
$\mathbf y_j^{\top}$ as rows. Let
$\mathbf P_m=\operatorname{softmax}(\mathbf Z_m)$.
For a minibatch of size $n$, the classifier gradients are
$\mathbf G_{W,m}=\frac{1}{n}(\mathbf P_m-\mathbf Y)^{\top}\mathbf H_m$
and
$\mathbf g_{b,m}=\frac{1}{n}\sum_{j=1}^{n}(\mathbf p_{m,j}-\mathbf y_j)$.
Flattening and concatenating them yields $\mathbf g^{p}_{m,t}$ without an
additional encoder forward or backward pass.
To distinguish random minibatch fluctuations from genuine temporal
changes, we model the probe gradient as:
\begin{equation}
\mathbf g^{p}_{m,t}
=
\mathbf s_{m,t}+\boldsymbol\epsilon_{m,t},
\qquad
\mathbf s_{m,t}
=
\mathbf s_{m,t-1}+\boldsymbol\omega_{m,t},
\label{eq:state}
\end{equation}
where $\boldsymbol\epsilon_{m,t}$ and $\boldsymbol\omega_{m,t}$
denote minibatch noise and temporal drift.
This decomposition separates stochastic fluctuations from temporal changes,
providing the basis for estimating the two dynamics independently. Accordingly, we denote the average per-dimension strengths of minibatch noise and temporal
drift by $R_{m,t}$ and $Q_{m,t}$, which are estimated online from the observed probe gradients.

\noindent\textbf{Minibatch Noise.}
At fixed model parameters, the discrepancy between two subsets of the same
minibatch mainly reflects stochastic sampling noise. We therefore split each
minibatch into two interleaved halves and estimate the full-minibatch noise
from their probe gradients $\mathbf g^{p,A}_{m,t}$ and
$\mathbf g^{p,B}_{m,t}$,
\begin{equation}
\widetilde R_{m,t}
=
\frac{1}{4}
\operatorname{mean}
\left[
\left(
\mathbf g^{p,A}_{m,t}
-
\mathbf g^{p,B}_{m,t}
\right)^2
\right].
\label{eq:R}
\end{equation}
Here, the factor \(1/4\) rescales the disagreement between two half-batch gradients to the noise level of the full-minibatch gradient, since each half contains only half of the samples.

\noindent\textbf{Temporal Drift.}
Unlike minibatch noise, temporal drift should capture the change of the
underlying gradient across consecutive iterations. However, the difference
between consecutive probe gradients also contains sampling noise. Let
$\Delta\mathbf g^{p}_{m,t}
=\mathbf g^{p}_{m,t}-\mathbf g^{p}_{m,t-1}$.
We therefore estimate the temporal drift as:
\begin{equation}
\widetilde Q_{m,t}
=
\max\!\left(
\frac{\|\Delta\mathbf g^{p}_{m,t}\|_2^2}{d_m}
-\widetilde R_{m,t}-\widetilde R_{m,t-1},
\;
\gamma_q\frac{\|\mathbf g^{p}_{m,t}\|_2^2}{d_m}
\right).
\label{eq:Q}
\end{equation}
The subtraction removes sampling-noise contributions from the inter-step
variation, allowing $\widetilde Q_{m,t}$ to better capture the underlying
temporal gradient change, while the second term prevents the estimate from
becoming excessively small. EMA smoothing then yields the stabilized
$R_{m,t}$ and $Q_{m,t}$.

\subsection{Centered Memory Controller}
\label{sec:memory_controller}

With $R_{m,t}$ and $Q_{m,t}$ estimated, we characterize the trade-off between
noise suppression and temporal tracking by the drift-to-noise ratio
$\rho_{m,t}=Q_{m,t}/R_{m,t}$. This ratio normalizes temporal variation by
noise, providing a scale-consistent measure of the tracking--smoothing
trade-off across modalities. We therefore use it as the control signal for
determining modality-specific gradient memory.

\noindent\textbf{Drift-to-Noise Gain.}
Although $\rho_{m,t}$ characterizes the relative gradient dynamics of each
modality, it does not provide a valid coefficient for temporal
memory. We therefore map this signal to a temporal gain using the
steady-state random-walk relation~\cite{kalman1960new},
\begin{equation}
\rho_{m,t}
=
\frac{K_{m,t}^{2}}{1-K_{m,t}},
\qquad
K_{m,t}
=
\frac{\sqrt{\rho_{m,t}^{\,2}+4\rho_{m,t}}-\rho_{m,t}}{2}.
\label{eq:kalman}
\end{equation}
Here, $K_{m,t}$ acts as the temporal gain that controls the response to
current-gradient information relative to accumulated history. This mapping
provides a model-derived basis for subsequent gradient-memory assignment
rather than heuristic momentum scaling.

\noindent\textbf{Cross-Modal Centering.}
The obtained $K_{m,t}$ reflects the tracking demand of each modality.
However, its absolute value is determined by the modality-specific
drift-to-noise ratio, and directly using these gains may collectively
shift the gradient memory away from the setting of the base optimizer.
We therefore preserve their relative differences while anchoring the
common memory level to the base momentum $\beta_0$.
Let $K_0=1-\beta_0$ and
$\ell_{m,t}=\operatorname{logit}(K_{m,t})$. We center the modal gains in
log-odds space as:
\begin{equation}
\widetilde{\ell}_{m,t}
=
\operatorname{logit}(K_0)
+
\lambda
\left(
\ell_{m,t}
-
\frac{1}{M}
\sum_{j=1}^{M}\ell_{j,t}
\right),
\label{eq:center}
\end{equation}
where the subtraction removes the cross-modal common shift and re-centers
the residual modal differences around $\operatorname{logit}(K_0)$. Thus,
the controller responds to relative modal dynamics while retaining the base
optimizer as the memory reference, with $\lambda$ controlling the strength
of cross-modal differentiation.

\noindent\textbf{Modal Memory Assignment.}
After cross-modal centering, $\widetilde{\ell}_{m,t}$ remains in log-odds
space and cannot be directly used as a momentum coefficient. We therefore
map it back to a bounded temporal gain and convert the gain into
modality-specific momentum,
\begin{equation}
\widetilde K_{m,t}
=
\operatorname{clip}
\left(
\sigma(\widetilde{\ell}_{m,t}),
K_{\min},
K_{\max}
\right),
\qquad
\beta_{m,t}
=
1-\widetilde K_{m,t}.
\label{eq:beta}
\end{equation}
Here, $\sigma$ maps the centered signal to $(0,1)$, while clipping restricts
the temporal gain to a stable range. Together with
$\beta_{m,t}=1-\widetilde K_{m,t}$, this converts the centered control signal
into a valid modality-specific momentum coefficient for adaptive gradient
memory.
\subsection{Modal Parameter Update}
\label{sec:modal_update}

With the modality-specific momentum $\beta_{m,t}$ obtained, we incorporate
the adaptive gradient memory into multimodal joint optimization. Since
$\beta_{m,t}$ varies throughout training, the conventional bias correction
for a fixed momentum coefficient is no longer applicable. We therefore
update and correct the first moment jointly as:
\begin{equation}
\mathbf m_{m,t}=\beta_{m,t}\mathbf m_{m,t-1}+(1-\beta_{m,t})\mathbf g^{\theta}_{m,t},
\;
\widetilde{\mathbf m}_{m,t}=\frac{\mathbf m_{m,t}}{1-\prod_{\tau=1}^{t}\beta_{m,\tau}},
\label{eq:modal_update}
\end{equation}
where $\mathbf g^{\theta}_{m,t}$ denotes the joint-training gradient of
modality $m$, and $\mathbf m_{m,t}$ accumulates its gradient history under
the time-varying momentum. The product $\prod_{\tau=1}^{t}\beta_{m,\tau}$
accounts for the complete momentum history, yielding the bias-corrected
first moment $\widetilde{\mathbf m}_{m,t}$. The corrected moment is then
passed to the base optimizer. Modality-specific parameters use
$\beta_{m,t}$, while shared and fusion parameters retain $\beta_0$,
allowing VCMM to adapt gradient memory without introducing an explicit
modality-specific learning-rate multiplier.

\section{Experiments}
\label{sec:experiments}

\begin{table}[!t]
\centering
\caption{Comparison with representative multimodal learning methods. 
The best results are highlighted in \textbf{bold}. 
The \underline{underline} denotes the second-best performance. 
Gray-shaded results indicate that multimodal performance is lower than that of the best unimodal approach.}
\label{tab:main}

\footnotesize 
\setlength{\tabcolsep}{2.5pt} 
\renewcommand{\arraystretch}{0.85}

\resizebox{\columnwidth}{!}{%
\begin{tabular}{l|cc|cc|cc|cc}
\toprule
\multirow{2}{*}{\textbf{Method}} 
& \multicolumn{2}{c|}{CREMA-D} 
& \multicolumn{2}{c|}{KSounds} 
& \multicolumn{2}{c|}{Twitter} 
& \multicolumn{2}{c}{NVGesture} \\
\cmidrule{2-9}
& Acc. & mAP & Acc. & mAP & Acc. & F1 & Acc. & F1 \\
\midrule

Unimodal-1 
& .6317 & .5879 & .5412 & .5669 & .5863 & .4333 & .7822 & .7833 \\

Unimodal-2 
& .4583 & .6861 & .5562 & .5837 & .7367 & .6849 & .7863 & .7865 \\

Unimodal-3 
& -- & -- & -- & -- & -- & -- & .8154 & .8183 \\

\midrule

Concat
& .6361 & \cellcolor{gray!20}.6841 
& .6455 & .7130 
& \cellcolor{gray!20}.7011 & \cellcolor{gray!20}.6386 
& .8237 & .8270 \\

\midrule

G-Blend~\cite{wang2020hard}
& .6465 & .7392 
& .6722 & .7274 
& \cellcolor{gray!20}.7309 & \cellcolor{gray!20}.6799 
& .8299 & .8305 \\

OGM~\cite{peng2022ogm}
& .6612 & .7372 
& .6582 & .7159 
& \cellcolor{gray!20}.7058 & \cellcolor{gray!20}.6435 
& -- & -- \\

PMR~\cite{fan2023pmr}
& .6659 & .7058 
& .6675 & .7274 
& \cellcolor{gray!20}.7357 & \cellcolor{gray!20}.6636 
& -- & -- \\

MLA~\cite{zhang2024mla}
& .7943 & .8572 
& .7004 & .7945 
& \cellcolor{gray!20}.7352 & \cellcolor{gray!20}.6713 
& .8340 & .8372 \\

ReconBoost~\cite{hua2024reconboost}
& .7557 & .8140 
& .6855 & .7662 
& .7442 & \cellcolor{gray!20}.6832 
& .8386 & .8434 \\

LFM~\cite{yang2024lfm}
& .8362 & .9006 
& .7253 & .7897 
& .7501 & \underline{.7057} 
& .8436 & .8468 \\

InfoReg~\cite{huang2025inforeg}
& .7498 & .8603 
& .7189 & \underline{.7986} 
& .7403 & \cellcolor{gray!20}.6847 
& -- & -- \\

AUG~\cite{jiang2025aug}
& \underline{.8515} & \underline{.9103} 
& .7263 & .7901 
& \underline{.7512} & .6962 
& \underline{.8501} & \underline{.8533} \\

DecAlign~\cite{qian2026decalign}
& .8426 & .9027 
& \underline{.7298} & .7926 
& .7396 & \cellcolor{gray!20}.6786 
& .8397 & .8429 \\

\midrule

\multirow{2}{*}{\textbf{VCMM}} 
& \textbf{.8562} & \textbf{.9148} 
& \textbf{.7534} & \textbf{.8218} 
& \textbf{.7570} & \textbf{.7098} 
& \textbf{.8532} & \textbf{.8558} \\[-0.8ex]

& {\tiny $\pm0.0024$} 
& {\tiny $\pm0.0036$} 
& {\tiny $\pm0.0025$} 
& {\tiny $\pm0.0037$} 
& {\tiny $\pm0.0018$} 
& {\tiny $\pm0.0023$} 
& {\tiny $\pm0.0018$} 
& {\tiny $\pm0.0011$} \\

\bottomrule
\end{tabular}%
}
\vspace{-1em}
\end{table}

\subsection{Experimental Setup}

\textbf{Datasets and Metrics.}
We evaluate VCMM on four multimodal benchmarks: CREMA-D~\cite{cao2014cremad}
and KSounds~\cite{arandjelovic2017look} for audio--visual learning,
Twitter~\cite{yu2019adapting} for image--text learning, and
NVGesture~\cite{molchanov2016online} with RGB, optical flow, and depth.
Following dataset-specific protocols, we report accuracy (Acc.), mAP, and
Macro-F1.

\noindent
\textbf{Implementation Details.}
Following OGM~\cite{peng2022ogm}, we use ResNet-18~\cite{he2016deep} for CREMA-D and KSounds, BERT~\cite{devlin2019bert} and ResNet-50 for Twitter, and pretrained I3D~\cite{carreira2017quo} for NVGesture. CREMA-D, KSounds, and NVGesture are optimized with Adam~\cite{kingma2015adam} using momentum $0.9$, weight decay $1\times10^{-4}$, and an initial learning rate of $1\times10^{-2}$, while Twitter uses Adam~\cite{kingma2015adam} with a learning rate of $2\times10^{-5}$ and weight decay $2\times10^{-4}$. The batch sizes are 64, 32, 32, and 2 for CREMA-D, KSounds, Twitter, and NVGesture, respectively. For VCMM, we set the base momentum to $\beta_0=0.9$, the statistics EMA coefficient to $a=0.95$, the adaptation strength to $\lambda=1$, and the drift floor to $\gamma_q=10^{-4}$, with the temporal gain clipped to $[0.01,0.30]$. The same VCMM hyperparameters are used across all datasets. All models are implemented with PyTorch and experiments are conducted on NVIDIA GeForce RTX 4090 GPUs.

\subsection{Main Results}

\begin{figure}[!t]
    \centering
    \includegraphics[width=\columnwidth]{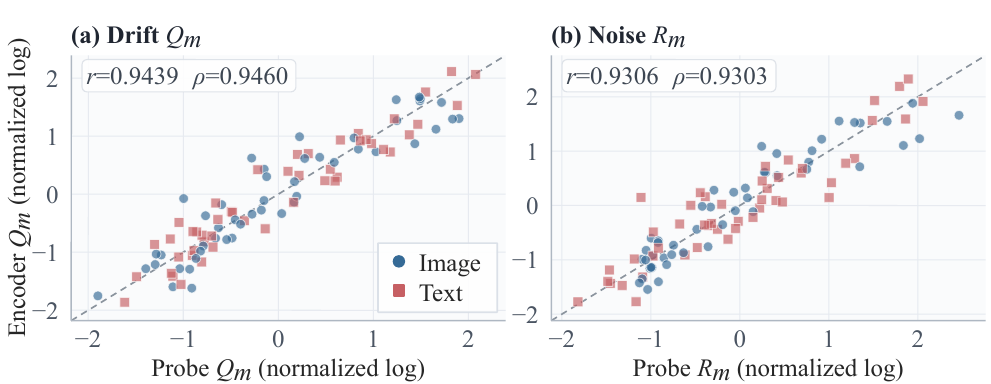}
\caption{
Agreement between the classifier probe and encoder-level estimates of
(a) temporal drift $Q_m$ and (b) minibatch noise $R_m$.
}
    \label{fig:probe_encoder_reference}
\end{figure}

\noindent\textbf{Comparison with Multimodal Baselines.}
Table~\ref{tab:main} compares VCMM with representative multimodal learning methods. Gray cells denote results below the best unimodal model. VCMM achieves the best performance on all eight metrics across the four datasets, showing consistent improvements across audio--visual, image--text, and trimodal settings.
The gains are particularly clear on KSounds, where VCMM achieves 75.34\% Acc and 82.18\% mAP, outperforming the strongest baselines by 2.36\% and 2.32\% respectively. VCMM also consistently improves the best competing results on CREMA-D, Twitter, and NVGesture. These results demonstrate that adapting gradient memory to modality-specific gradient dynamics effectively improves multimodal joint optimization.
\vspace{-1em}

\subsection{Ablation Study}
\label{sec:ablation}

We conduct ablation studies on KSounds and Twitter with the backbone,
optimizer, and training configuration fixed across all variants.
Shared $\beta$ uses $\beta=0.9$ for all modalities and fixed asymmetric
selects modality-specific momentum coefficients from
$\{0.70,0.80,0.90,0.95,0.99\}$ on the validation set and keeps them
fixed throughout training. Frozen VCMM runs VCMM during the initial
10\% of training and then fixes each modal momentum to its average
over the final part of this period.
As shown in Table~\ref{tab:ablation}, fixed asymmetric momentum
consistently improves over shared $\beta$, indicating the benefit of
modality-specific gradient memory. Full VCMM further outperforms
Frozen VCMM by 7.21\% and 1.58\% on KSounds and Twitter,
respectively, demonstrating the importance of time-varying adaptation.
Removing noise subtraction, log-odds centering, or exact bias correction
consistently degrades performance. Noise subtraction has the largest
impact on KSounds, while log-odds centering contributes most on Twitter,
highlighting the importance of reliable dynamics estimation and
cross-modal calibration.

\subsection{Further Analysis}
\label{sec:further_analysis}

\begin{figure}[!t]
    \centering
    \includegraphics[width=\columnwidth]{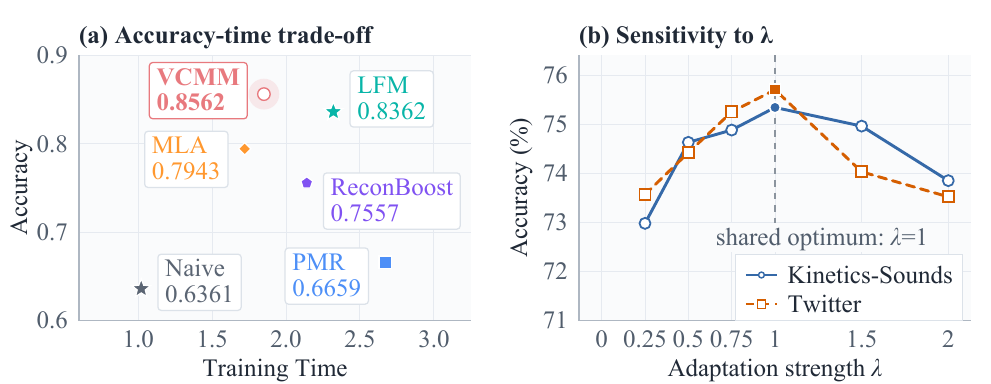}
    \caption{(a) Accuracy--training time trade-off on CREMA-D. (b) Sensitivity to the adaptation strength $\lambda$ on KSounds and Twitter.}
    \label{fig:accuracy_time_lambda_sensitivity}
\end{figure}

\begin{table}[!t]
\centering

\begin{minipage}[t]{0.48\columnwidth}
\centering
\vspace{-1em}
\caption{Ablation on KSounds and Twitter.}
\label{tab:ablation}
\vspace{1mm}
\scriptsize
\setlength{\tabcolsep}{1.5pt}
\renewcommand{\arraystretch}{0.95}
\begin{tabular}{@{}lcc@{}}
\toprule
\textbf{Variant} & \textbf{KSounds} & \textbf{Twitter} \\
\midrule
Shared $\beta$         & 0.6455 & 0.7011 \\
Fixed asymmetric       & 0.6773 & 0.7396 \\
Frozen VCMM            & 0.6813 & 0.7412 \\
w/o noise subtraction  & 0.6935 & 0.7499 \\
w/o log-odds centering & 0.7070 & 0.7452 \\
w/o exact bias corr.   & 0.7208 & 0.7513 \\
\midrule
\textbf{Full VCMM} & \textbf{0.7534} & \textbf{0.7570} \\
\bottomrule
\end{tabular}
\end{minipage}\hfill
\begin{minipage}[t]{0.48\columnwidth}
\centering
\vspace{-1em}
\caption{Training memory cost on CREMA-D.}
\vspace{3pt}
\scalebox{0.87}{
\label{tab:training_cost}
\scriptsize
\setlength{\tabcolsep}{2pt}
\renewcommand{\arraystretch}{0.95}

\resizebox{\linewidth}{!}{%
\begin{tabular}{@{}lcc@{}}
\toprule
\multirow{2}{*}{\textbf{Method}}
& \textbf{Alloc.}
& \textbf{Overhead} \\
& \textbf{(GiB)}
& \textbf{(\%)} \\
\midrule
Concat                          & 8.89  & Base  \\
MLA~\cite{zhang2024mla}         & 11.13 & +25.2 \\
PMR~\cite{fan2023pmr}           & 12.38 & +39.3 \\
LFM~\cite{yang2024lfm}          & 10.88 & +22.3 \\
ReconBoost~\cite{hua2024reconboost}
                                & 11.37 & +27.9 \\
\midrule
\textbf{VCMM}
& \textbf{10.10}
& \textbf{+13.6} \\
\bottomrule
\end{tabular}%
}
}
\end{minipage}

\end{table}

\begin{figure}[!t]
    \centering
    \includegraphics[width=\columnwidth]{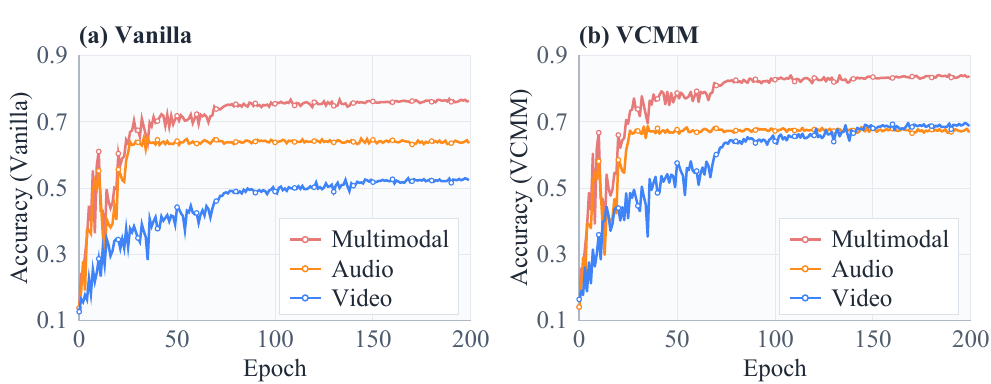}
\caption{Modality-wise validation accuracy during joint training. (a) Vanilla. (b) VCMM.}
    \label{fig:converge}
\end{figure}
\noindent\textbf{Computational Efficiency.}
As illustrated in figure~\ref{fig:accuracy_time_lambda_sensitivity} (a) and table~\ref{tab:training_cost}, VCMM achieves 85.62\% accuracy
with only 13.6\% memory overhead over Concat, while maintaining competitive
training time. These results demonstrate a favorable accuracy--efficiency trade-off,
enabled by the lightweight probe without additional encoder passes.

\noindent\textbf{Alleviation of Modality Imbalance.}
As shown in Figure~\ref{fig:converge}, vanilla training exhibits a clear
modality gap, whereas VCMM enables video to continue improving while
maintaining audio performance. The narrowing gap indicates that adaptive
gradient memory sustains optimization of the slower modality without degrading
the faster one, thereby alleviating modality imbalance.

\noindent\textbf{Agreement with Encoder-Level Dynamics.}
As shown in Fig.~\ref{fig:probe_encoder_reference}, the classifier probe
closely matches the encoder-level estimates of temporal drift $Q_m$ and
minibatch noise $R_m$, with Pearson correlations of $0.9439$ and $0.9306$.
This validates the lightweight probe as an efficient surrogate for
encoder-level gradient dynamics.

\noindent\textbf{Effect of Adaptation Strength $\lambda$.}
Figure~\ref{fig:accuracy_time_lambda_sensitivity} (b) shows similar trends
on KSounds and Twitter, with the best performance around $\lambda=1$.
Moderate cross-modal differentiation benefits adaptive gradient memory,
whereas overly large $\lambda$ can overemphasize modal differences and
degrade performance. The consistent trend across both datasets indicates
robustness around $\lambda=1$.

\section{Conclusion}
In this work, we revisit modality imbalance from the perspective
of gradient memory and show that different modalities exhibit
distinct and time-varying gradient dynamics, making shared
momentum potentially suboptimal for multimodal joint training.
Based on this observation, we propose \textbf{V}ariance-\textbf{C}alibrated
\textbf{M}omentu\textbf{M} (\textbf{VCMM}), which adapts modality-specific gradient memory
according to online estimates of minibatch noise and temporal drift
without explicit learning-rate scaling. Experiments across
four multimodal benchmarks demonstrate consistent improvements
over joint training and competitive baselines with modest
computational overhead, while effectively alleviating modality imbalance.

\clearpage
\section{COMPLIANCE WITH ETHICAL STANDARDS}
This study uses only publicly available datasets and does not involve
the collection of new human or animal subject data. Therefore, ethical
approval was not required.
\bibliographystyle{IEEEbib}
\bibliography{refs}

\end{document}